\documentclass[twoside]{article}

\usepackage{PRIMEarxiv}

\usepackage[utf8]{inputenc} 
\usepackage[T1]{fontenc}    
\usepackage{hyperref}       
\usepackage{url}            
\usepackage{booktabs}       
\usepackage{amsfonts}       
\usepackage{nicefrac}       
\usepackage{microtype}      
\usepackage{lipsum}
\usepackage{fancyhdr}       
\usepackage{graphicx}       
\usepackage{amsmath}
\usepackage{caption}
\usepackage{subcaption}
\usepackage{mdframed}
\usepackage[backend=biber]{biblatex}
\graphicspath{{media/}}     

\title{
What makes for good morphology representations for spatial omics?
}

\author{
  Eduard Chelebian, Christophe Avenel, Carolina Wählby \\
  Department of Information Technology and SciLifeLab \\
  Uppsala University \\
  Uppsala, Sweden\\
  \texttt{\{eduard.chelebian, christophe.avenel, carolina.wahlby\}@it.uu.se} \\
}

\begin{document}
\maketitle

\begin{abstract}
Spatial omics has transformed our understanding of tissue architecture by preserving spatial context of gene expression patterns. Simultaneously, advances in imaging AI have enabled extraction of morphological features describing the tissue. The intersection of spatial omics and imaging AI presents opportunities for a more holistic understanding. In this review we introduce a framework for categorizing spatial omics-morphology combination methods, focusing on how morphological features can be translated or integrated into spatial omics analyses. By translation we mean finding morphological features that spatially correlate with gene expression patterns with the purpose of predicting gene expression. Such features can be used to generate super-resolution gene expression maps or infer genetic information from clinical H\&E-stained samples. By integration we mean finding morphological features that spatially complement gene expression patterns with the purpose of enriching information. Such features can be used to define spatial domains, especially where gene expression has preceded morphological changes and where morphology remains after gene expression. We discuss learning strategies and directions for further development of the field. 
\end{abstract}


\section{Introduction}

Before the advent of spatial omics technologies, single-cell omics relied on tissue dissociation which resulted in the loss of spatial information for understanding cellular environments and interactions \cite{longo2021integrating, rao2021exploring, bressan2023dawn, park2022spatial}. Preserving the spatial context of cells provides a more comprehensive view of cellular heterogeneity and tissue architecture, leading to deeper insights into the molecular landscape of disease \cite{seferbekova2023spatial, zhao2022spatial, zhou2023spatial}. The early analysis methods in spatial omics naturally evolved from established analysis techniques by the single-cell community \cite{vandereyken2023methods}. However, this reliance often led to an oversight of the spatial and morphological relationships present within tissues. 

Parallelly, the artificial intelligence (AI) imaging community have significantly enhanced our ability to analyze and interpret image data. Deep learning algorithms, convolutional neural networks (CNN) and image segmentation techniques have enabled for more precise analysis of biological tissues \cite{cui2021artificial, song2023artificial}. Spatial omics, which captures spatially-resolved data and usually includes imaging data, naturally intersects with such technologies. By combining the gene expression information from spatial omics and morphological features from imaging, we can achieve a more holistic understanding of tissue architecture \cite{lu2021integrative}. Nonetheless, there are some challenges due in part to the disconnect between the expertise of both communities. It is inherently hard to simultaneously utilize two different data sources of such high dimensionality as spatial omics and imaging data. While obtaining features from spatial omics is relatively established and interpretable, the patterns often directly representing local gene expression, extracting relevant morphological features from images requires more understanding and careful validation.

Combining advancements in imaging AI into spatial omics would benefit from a structured approach to leverage these combined data sources effectively. Previous reviews and benchmarks have explored separate aspects of the combination \cite{chan2023benchmarking, yuan2024benchmarking} or more generally the applications of AI for spatial omics \cite{li2022emerging, zahedi2024deep}. This Review introduces a comprehensive framework to understand such combination methods by analyzing the distinct ways morphological features can be utilized. Image morphology can either be \textit{translated} into features that correlate with spatial omics, taking advantage of the ease and cost-effectiveness of acquiring it; or \textit{integrated} with spatial omics to provide a richer description of the sample. These approaches are inherently conflicting: translation focuses on gene-correlated features, while integration benefits from complementary information for a fuller understanding. This Review aims to clarify these distinctions to optimize the use of morphological and molecular data together.

\begin{figure}[ht]
     \centering
     \begin{subfigure}[b]{\textwidth}
         \centering
        \includegraphics[clip, trim=0cm 7cm 6cm 0cm, width=.8\linewidth]{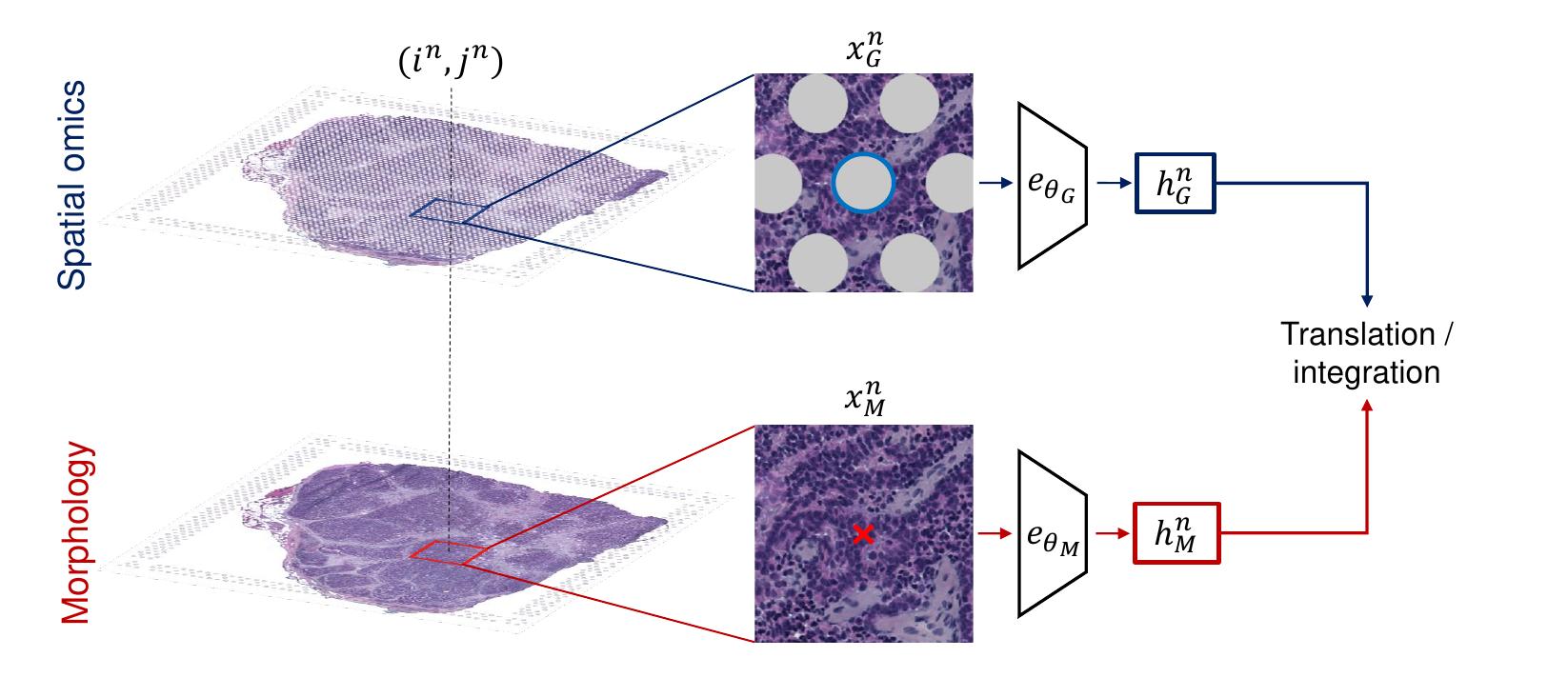}
         \caption{}
         \label{fig:preliminary}
     \end{subfigure}
     \vfill
     \begin{subfigure}[b]{\textwidth}
         \centering
    \includegraphics[clip, trim=.2cm 5.5cm 0cm 0cm, width=1\linewidth]{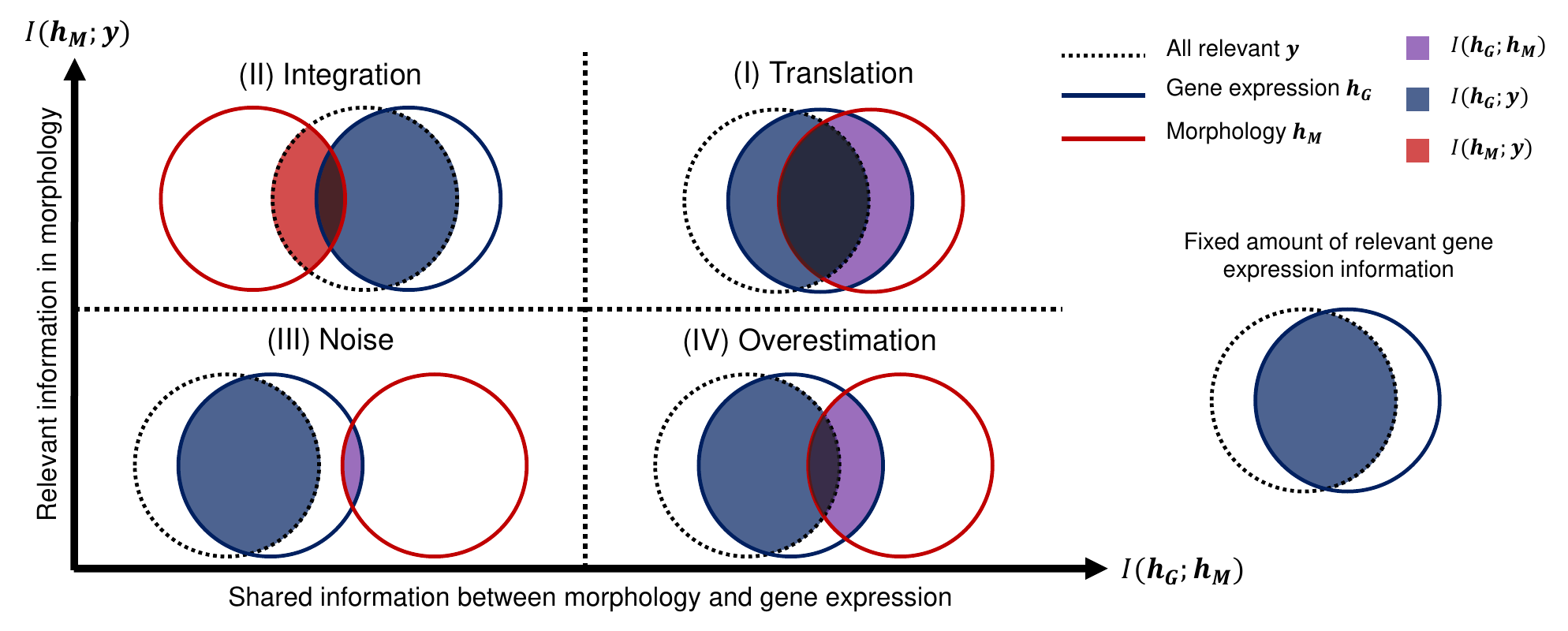}
         \caption{}
         \label{fig:intuition}
     \end{subfigure}
    \caption{(a) Feature extraction from spatial omics and morphology on 10X Visium spatial transcriptomics and H\&E \cite{erickson2022spatially}. We use the $n$th spot in Visium $x_{G}^n$ as the center to extract the $n$th image patch $x_{M}^n$. The modality specific encoders $e_{\theta_{G}}$ and $e_{\theta_M}$ will output the molecular feature vector $h_{G}^n$ and the morphological feature vector $h_{M}^n$ for position $(i^n, j^n)$. (b) Intuition of the framework. By assuming the amount of relevant spatial omics information is fixed (solid blue area), we get four different scenarios depending on the morphological features. We express these four scenarios by the quadrants formed when presenting the relevance of the morphological features as y-axis and their shared information with the spatial omics features as x-axis.}
    \label{fig:main}
\end{figure}

\section{Translation-integration framework for joint analysis of morphology and spatial omics}

Spatial omics, whether imaging-based or sequencing-based, in practice yields a grid of spatial positions together with gene expression information \cite{rao2021exploring}. The process for obtaining the gene expression is quite standardized for the different modalities of spatial omics. What users receive from sequecing-based solutions such as 10X Visium, slide-seq \cite{rodriques2019slide} or stereo-seq \cite{chen2022spatiotemporal} is a matrix of spots or cells representing gene expression together with their spatial coordinates, ready for downstream analysis. Similarly, imaging-based solutions, regardless if they are \textit{in situ} hybridization-based \cite{chen2015spatially, eng2019transcriptome}, or \textit{in situ} sequencing-based \cite{ke2013situ, larsson2010situ},  yield a stack of images with signals that need to be detected and decoded to retrieve the location of specific genes. Even if the specific methods to do so can vary, the general workflow is usually the same. The individual detections are then aggregated per cell to, again, construct a matrix of gene expression together with their spatial coordinates. The gene expression from spatial omics approaches on their own have shown their biological relevance on countless works \cite{seferbekova2023spatial, zhao2022spatial, zhou2023spatial}. But  this information is fixed and, once the appropriate analyses are in place, it is not possible to tweak the gene expression to obtain more relevant information.  

Many spatial omics approaches are paired with some kind of morphological imaging, usually stained with hematoxilyn-eosin (H\&E) or DAPI. In order to combine the gene expression with the images, it is common to pair them by extracting patches using the gene expression coordinates as patch centers, as shown in Figure \ref{fig:main}a. The process of obtaining meaningful characteristics from images is known as feature extraction. Traditionally, features were extracted using hand-crafted algorithms with known specific outcomes in mind. However, learning-based algorithms have proven to capture more powerful in a variety of medical tasks \cite{cui2021artificial, song2023artificial}. Contrary to gene expression from spatial omics, the features from images can be learned to be task-specific. For instance, the morphological features extracted for performing cancer grading can be very different from the morphological features for detecting mitosis for the same piece of tissue \cite{janowczyk2016deep}. Thus, the framework that we propose focuses on the design choices regarding the learned morphological features, their relevance and their interaction with gene expression.

The features we extract from morphological images should clearly be as relevant as possible, which is in itself a noteworthy task. Further, relevance can usually only be measured in these cases by assessing the performance of the features in a downstream task. But relevance in itself is not sufficient for a joint analysis of morphology and spatial omics. The morphological features need to have specific synergies with gene expression patterns in order to consider them appropriate for different joint analyses. If we construct a plot with these two dimensions, relevant information in the morphology features on the y-axis and shared information between morphology and gene expression on the x-axis, we obtain the four quadrants depicted in Figure \ref{fig:main}b. The two first quadrants (top: translation and integration) represent scenarios in which features  can be used for joint analysis with spatial omics, while the two last quadrants (bottom: noise and overestimation) represent scenarios in which features should not be used for joints analyses with spatial omics:

\begin{itemize}

    \item[I] \textbf{Translation}. In this scenario, the morphological features contain a high amount of relevant information and share a high amount of information with gene expression. These features include the same information as the task-relevant part of the gene expression. This scenario is ideal for gene expression prediction to, for instance, generate super-resolution maps or infer genetic information from clinical H\&E-stained samples without incurring in the costs. These features are typically obtained by specifically training deep learning models for that purpose. 

    \item[II] \textbf{Integration}. In this scenario, the morphological features contain a high amount of relevant information but do not share information with gene expression. These features contain relevant information that has not been captured by the gene expression. This scenario is ideal for spatial domain identification due to the time uncoupling between changes in gene expression and its effect on morphology. These features need to be more general as they should complement the information in the gene expression and not be redundant.

    \item[III] \textbf{Noise}. In this scenario, the morphological features do not contain relevant information nor they share information with gene expression. These features capture non-relevant variations such as sample-to-sample or staining variations which cannot be leveraged in a joint analysis with spatial omics as they will confuse the analysis.
    
    \item[IV] \textbf{Overestimation}. In this scenario, the morphological features do not contain relevant information but they do share information with gene expression. These features cover the gene expression that is not task-relevant and thus can lead to an overestimation of the predictive power in, for instance, translation tasks. These can be housekeeping or constitutive genes that may not be relevant for the specific downstream task.  
    
\end{itemize}

This framework already enables rethinking the way we design the morphological features (e.g. by training a neural network) for the different tasks and whether the joint analysis is even valuable or not. For instance, if we have morphological features that are highly correlated with gene expression, integration is a meaningless, if not harmful, as we would be including redundant information in our model. Equally, doing translation with morphological features that do not share information with gene expression would not be effective.  

The formal definition of this framework is presented in Box 1 and we present a practical example in Box 2. 

\clearpage

\begin{mdframed}

\subsection*{Box 1 -- Formal framework definition}

Let $\mathcal{D}  = \{ (x_{G}^1,   x_{M}^1, y^1), ..., (x_{G}^N, x_{M}^N, y^N)\}$ be the paired multi-modal dataset of two modalities (gene expression $G$ and morphology $M$) and $N$ number of data points with coordinates $\{(i^1,j^1), ..., (i^N,j^N)\}$. $y^n$ represents the potential task-relevant information that is contained in each area.

The spatial omics workflow usually consists on processing the data until we obtain, per each spatial position, a feature corresponding to the point- or aggregate-gene expression in the area. For the sake of notation, we represent these features as the result of encoding the spatial omics data with a gene expression-specific encoder $\boldsymbol{h_{G}} = e_{\theta_{G}}(\boldsymbol{x_{G}})$. Parallelly, we can represent the morphological features obtained from the paired images by an image-specific encoder  $\boldsymbol{h_{M}} = e_{\theta_{M}}(\boldsymbol{x_{M}})$. 

Finally, based on the ideas of minimal sufficient statistics \cite{soatto2014visual} and information bottleneck theory \cite{alemi2016deep, tishby2000information} proposed by Tian \textit{et al.} \cite{tian2020makes}, we define $I(\boldsymbol{a}; \boldsymbol{b})$ as the shared information between $\boldsymbol{a}$ and $\boldsymbol{b}$.

As the spatial omics workflow is quite established by the bioinformatics community, we can assume that the amount of task-relevant information contained in the gene expression is fixed and unknown, but not negligible:

\begin{equation}
    I ( \boldsymbol{h_{G}} ; \boldsymbol{y} ) > 0
\end{equation}

The only thing we can control then is how we train our feature extractor for the morphology data. Thus, we assume that we can only control $e_{\theta_{M}}$, the morphology-specific encoder, and the only relationship we can capture is $I ( \boldsymbol{h_{G}} ; \boldsymbol{h_{M}} )$.

Undoubtedly, we always want the morphological features to be as informative as possible:

\begin{equation}
    \max_{\theta_{M}} I ( \boldsymbol{h_{M}} ; \boldsymbol{y} )
\end{equation}

But depending on what we are using these features for, they will interact differently with gene expression. This creates four different scenarios depending on the interaction between the morphology-spatial omics and morphology-relevant spaces. We interpret these four scenarios by the four quadrants, as depicted in Figure~\ref{fig:main}b.

\textbf{Translation}. We want to ensure that the morphology contains the maximum task-relevant information. For this we can use the information included in gene expression as a proxy for relevance. This would require to maximize the shared information between both modalities while maintaining the relevance of morphology:

\begin{equation}
    \max_{\theta_M} I ( \boldsymbol{h_{G}} ; \boldsymbol{h_{M}} )
\end{equation}

so that the information shared between modalities is relevant, achieving the sweet point \cite{tian2020makes}:

\begin{equation}
I (\boldsymbol{h_{G}} ; \boldsymbol{h_{M}} ) = I (\boldsymbol{h_{G}} ; \boldsymbol{y}) = I (\boldsymbol{h_{M}} ; \boldsymbol{y} )
\end{equation}

\textbf{Integration}. We want to ensure that the fusion, by a fusion module $f_\psi$, between the modalities carries more information than the individual modalities, minimizing redundancies between modalities, without adding nuisance information.

\begin{equation}
\min_{\theta_M}  I ( \boldsymbol{h_{G}} ; \boldsymbol{h_{M}} )
\end{equation}

so that the addition of morphology to gene expression carries more task relevant information than gene expression alone:

\begin{equation}
I ( f_\psi ( \boldsymbol{h_{G}} , \boldsymbol{h_{M}} ) ; \boldsymbol{y}) > I (\boldsymbol{h_{G}} ; \boldsymbol{y}) 
\end{equation}

\end{mdframed}

\clearpage

\begin{mdframed}

\subsection*{Box 2 -- Practical example for the framework}

We want to visualize these four quadrants in practice. For this, we need a proxy for measuring the relevance of the morphological features and another for measuring the shared information with gene expression. For relevance, a simple approach would be to use pathologist annotations on an H\&E slide as the ground truth and with a mutual information-based feature selection approach, see the information the morphological features share with the annotations. Thus, the higher in the y-axis the more descriptive the feature is towards pathologist annotations. For shared information with gene expression, we could use the absolute Pearson correlation of the feature with its most correlated gene. This would show if there is a morphological feature that shows the same spatial patterns expression of a specific gene. We can compare this plot for two common feature extraction scenarios, using a model pretrained on ImageNet \cite{deng2009imagenet} and using the same model pretrained in a self-supervised way on millions of H\&E slides \cite{ciga2022self}. What we can immediately see from the plot below is a shift of the centroid of the distribution further from the third quadrant (noise) and towards the first quadrant (translation). So, from this simple analysis, we can conclude for instance that self-supervised pretraining on histology images would be more appropriate for a translation task than pretraining on natural images.

{
    \centering
    \includegraphics[clip, width=.7\linewidth]{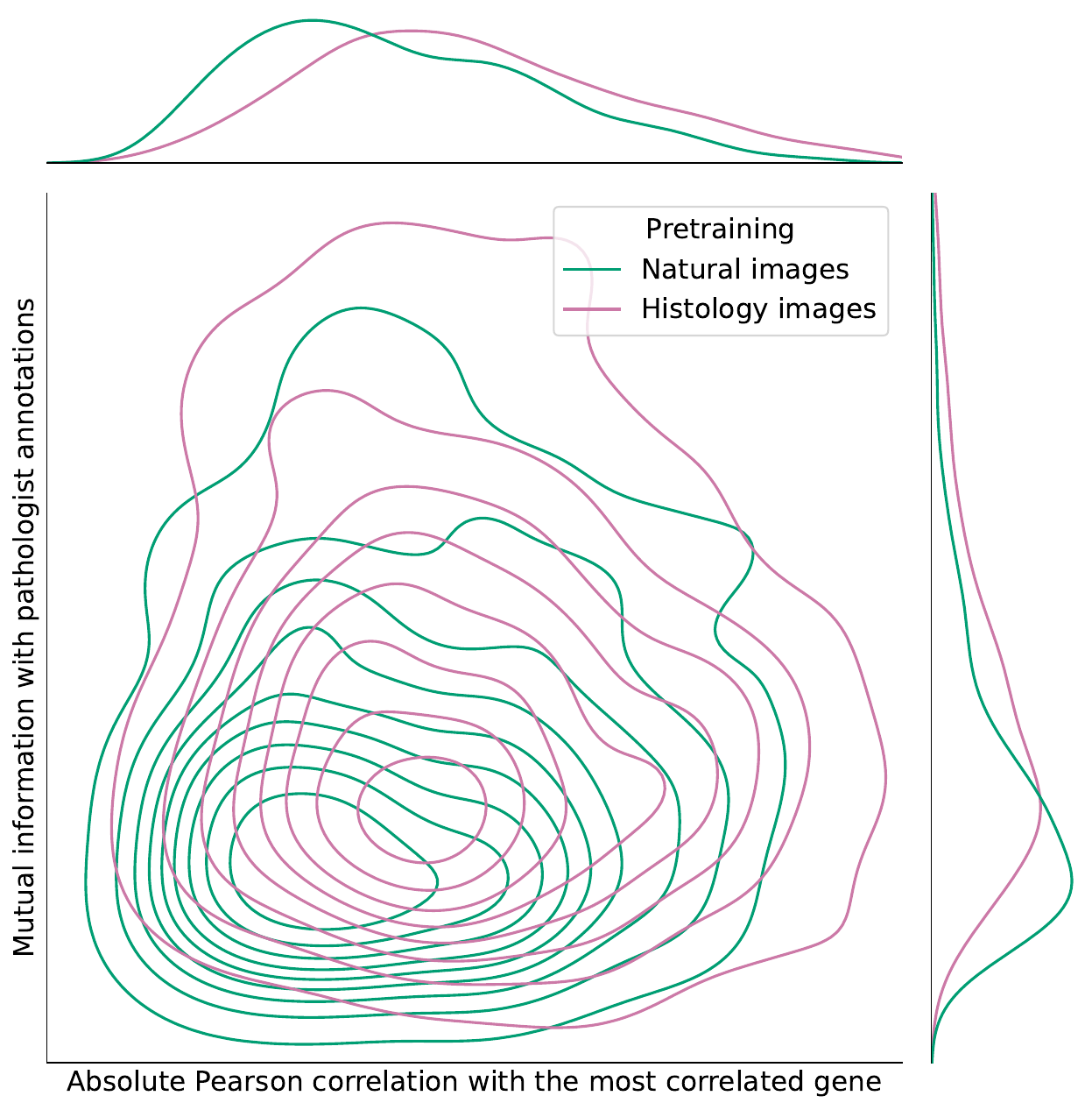}
    \par
}

\end{mdframed}

\section{Morphology translation for gene expression prediction}

\begin{figure}[ht]
    \centering
    {\includegraphics[clip, trim=0cm 7cm 20.5cm 0cm, width=.5\linewidth]{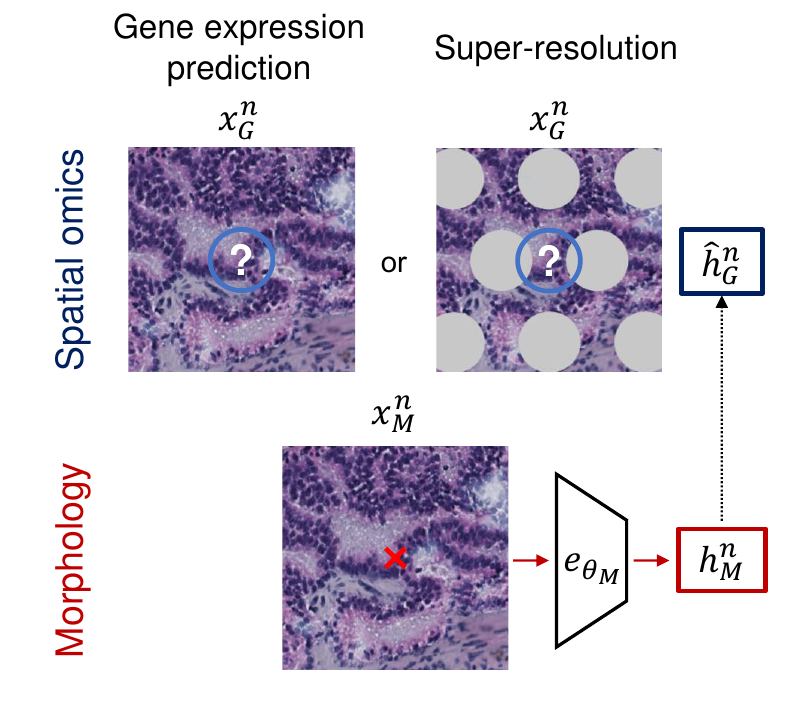}}
    \caption{The most common task involving morphology translation is the prediction of gene expression. This can be done either by inferring the gene expression in a new sample or by imputing the gene expression in between areas containing gene expression.}
    \label{fig:translation}
\end{figure}

Morphology translation involves identifying morphological features that spatially correlate with gene expression patterns. The primary application in this scenario is gene expression prediction. The goal is to learn morphological features that are highly correlated with gene expression so that these features can be used independently of gene expression data, either on new samples or in regions of a sample where spatial gene expression data is unavailable or sparse (see Figure \ref{fig:translation}).

Note that, within our framework, we specifically refer to the relationship of morphological images with spatial omics. Previous work has also tried to predict spatial gene expression using non-spatial sources \cite{mondol2023hist2rna, comiter2023inference, wang2021predicting, weitz2022transcriptome}. 

Using spatial gene data, the most common application has been to try and predict 10X Visium and other sequencing-based spatial trascriptomics expression from H\&E images and we will start this section with an historical perspective of such methods.

To our knowledge, this field began to take shape with the development of ST-Net \cite{he2020integrating}, paradoxically entitled "\textit{Integrating} spatial gene expression and breast tumor morphology via deep learning" which, per our definition, was a translation application rather than integration. The authors identified known breast cancer biomarkers and, even though the correlation values were moderate, they showed that the translation was relevant. 

The still unpublished HisToGene \cite{pang2021leveraging} is one of the first works available introducing the concept of super-resolution for spatial transcriptomics. The authors translate the gene expression to a dense map of the whole image, theoretically being able to achieve pixel-level expression. Monjo \textit{et al.} trained DeepSpaCE \cite{monjo2022efficient} to achieve super-resolution gene expression of areas between the spots. Additionally, they performed translation on consecutive sections in a semi-supervised way.

As the field matured, researchers began to incorporate more sophisticated comparisons. The authors of BrST-Net \cite{rahaman2023breast} extended ST-Net by analyzing various feature extracting techniques, while SEPAL \cite{mejia2023sepal} evaluated different approaches for predicting gene expression: global for the whole-image, patch-based local and spatially-informed, introducing the need of context in the prediction. These advances led to the developement of other graph-based methods that considered the local context such as THItoGene \cite{jia2024thitogene}, EGN \cite{yang2024spatial} or ErwaNet \cite{chen2024edge}.

Advances in the AI imaging community enabled leveraging the multi-resolution nature of H\&E images. M2ORT \cite{wang2024m2ort} and TRIPLEX \cite{chung2024accurate} proposed multi-scale feautre extractors, achieveing notable correlation improvements accross various datasets. And, recently, the authors of BLEEP \cite{xie2024spatially} explicitly trained a bi-modal embedding-based framework for spatial transcriptomics prediction from H\&E. 

\subsection{Learning features correlated with gene expression}

The general task for morphology translation methods is to input an image patch and output the gene expression patterns in that area. This task requires a series of design choices to effectively capture and predict the complex relationships between morphological features and gene expression. 

\subsubsection{Selection of training genes}

Selecting which genes the model should learn is crucial because sequencing-based spatial transcriptomics methods typically cover the entire transcriptome, encompassing nearly 20,000 genes. However, many of these genes do not exhibit spatial patterns or have low expression levels. To ensure the learned features are meaningful and fall within relevant quadrants of our framework, careful gene selection is necessary.

Three main approaches, set by the first works, have been established for gene selection, which have the same goal:

\textbf{Highest mean expression}. This method involves selecting genes that have the highest average expression levels across the dataset. The rationale is that genes with higher mean expression are likely to be more robust and exhibit clearer spatial patterns. ST-Net and BrST-Net utilized this approach to focus on the most consistently expressed genes. 

\textbf{Highly variable genes}. This strategy targets genes with high variability in their expression levels across different spatial locations. Genes with significant expression variability are more likely to capture meaningful spatial differences. Methods like HisToGene, THItoGene, M2ORT, TRIPLEX, EGC or ErwaNet have adopted this approach, aiming to enhance the model’s ability to learn diverse and informative patterns.

\textbf{Manual selection}. In this approach, researchers manually select genes based on prior biological knowledge or specific research goals. This method allows for the inclusion of genes known to be relevant to particular conditions or tissues, ensuring that the selected genes are biologically significant. DeepSpaCE or BLEEP used manual selection to tailor their model to specific biomarkers of interest.

These approaches highlight that current methods are not yet capable of predicting the entire transcriptome due to the excessive noise present in such highly dimensional data.

\subsubsection{Training regimes}

Once we have selected our training genes, the next crucial step is to choose how to learn morphological features that correlate with these genes. The general approach involves a straightforward deep learning regression pipeline: input -- model -- output. Here, the input consists of image patches, and the output is the gene expression patterns for those patches. The main variable in this setup is the type of model used.

Early methods in this field leaned heavily on convolutional neural networks (CNNs) due to their proven effectiveness in image analysis. These models excel in capturing spatial hierarchies through their convolutional layers. For instance, ST-Net employed DenseNet-121 \cite{huang2017densely}. DeepSpaCE, another pioneering method, utilized the simpler VGG16 \cite{simonyan2014very} which offered a straightforward yet powerful tool for feature extraction. BrST-Net expanded on this approach by comparing various CNNs architectures and ultimately identified EfficientNet \cite{tan2019efficientnet} as the top performer.

One significant shift was towards Transformer-based models, which are adept at capturing long-range dependencies in data. HisToGene was an early adopter of the Vision Transformer (ViT) \cite{dosovitskiy2020image}, which applies the transformer architecture directly to image patches, treating them as sequences of tokens. This ability to capture global context more effectively than traditional CNNs was beneficial for understanding complex spatial patterns in gene expression data.

To further enhance the understanding of spatial organization, researchers introduced Graph Neural Networks (GNNs) \cite{wu2020comprehensive}. SEPAL combined the strengths of ViTs for obtaining image embeddings with GNNs to learn spatial patterns. This approach leveraged the global feature capturing of transformers and the spatial relationship modeling of GNNs, providing a comprehensive framework for morphology translation. Similarly, THItoGene used Efficient-Capsule Networks \cite{mazzia2021efficient}, designed to better capture spatial hierarchies, to generate embeddings and combined ViTs and GNNs for reconstructing gene expression. 

The pyramidal nature of histopathological images, with features at multiple scales, led to the development of multi-level models. M2ORT, for example, used a hierarchical ViT \cite{chen2022scaling} to accommodate the multi-scale nature of these images. By processing images at different scales, the hierarchical ViT could capture fine-grained details as well as broader contextual information, making it particularly effective for gene expression prediction. TRIPLEX adopted a similar approach with a multi-resolution architecture based on ResNet-18 \cite{he2016deep}. 

Recent advancements introduced bi-modal embedding-based frameworks, which aim to create joint representations of images and gene expression data. BLEEP was the first to explicitly train such a framework, similar to CLIP \cite{radford2021learning}. 

\subsubsection{Training splits}

The choice of training splits is critical for the validation of the generalization abilities of deep learning models, especially given the typically small size of these datasets.

A common approach is leave-one-out validation. It involves using each sample as the test set, one at a time, while the remainder are used as a training set. While it maximizes the use of the dataset, it provides limited view on the generalization of the model. ST-Net, HisToGene, and THitoGene have employed leave-one-out validation to assess their models.

Another approach for small datasets is k-fold cross validation, where the dataset is divided in $k$ subsets and, one at a time, are used as test and the remaining for training. This ensures that the performance is at least averaged accross different partitions and it is the approach of BrST-Net, TRIPLEX, EGN and ErwaNet.

Finally, methods such as SEPAL, BLEEP and M2ORT use traditional train--validation--test partitions which can suffer from bad splitting choices but show generalization ability on independent samples. BrST-Net and TRIPLEX validated their method using leave-one-out and k-fold cross validation but additionally tested on an independent sample.

\subsection{Other tasks involving morphology translation}

Although super-resolution can be obtained by dense prediction of gene expression, some authors have proposed other methods to achieve this. The earliest work published actually achieving pixel-level super-resolution spatial transcriptomics from H\&E is, to our knowledge, XFuse \cite{bergenstraahle2022super}. At the time, super-resolution lacked a ground-truth benchmark for comparison, prompting the authors to demonstrate the robustness and potential of their method by comparing their results with \textit{in situ} hybridization. iStar by Zhang \textit{et al.} \cite{zhang2024inferring} is another super-resolution method that aims at predicting gene expression at super-resolution level. The newer Xenium technology \cite{janesick2023high} with higher gene expression resolution, enabled a quantitative comparison with XFuse, showing that their method enabled predictions that were closer to the ground truth. 

One of the few instances where translation by deep learning was made from DAPI images instead of H\&E was demonstrated by us \cite{andersson2020transcriptome}. We utilized imaging-based spatial omics to annotate marker genes, training CNNs to classify tissue morphology. This adaptation indicates the versatility of deep learning methods in handling different imaging types for morphology translation. DAPI imaging data is often neglected but holds a lot of potential as shown in \cite{chelebian2024self}.

We also explored the correlation of latent features in different networks with gene expression \cite{chelebian2021morphological}. We discovered that networks trained for cancer classification tasks \cite{strom2020artificial} already contained features correlating with genes associated with prostate cancer \cite{erickson2022spatially}. Building on this, we introduced MHAST, a framework that employs self-supervised features to guide the re-assignment of deconvolved spatial transcriptomics \cite{chelebian2024learned}. Our validation on Tangram \cite{biancalani2021deep} demonstrated that MHAST could retrieve more accurate cell instances compared to original random allocations, showcasing its potential for achieving cell-level resolution.

More recently, Gao \textit{et al.} presented IGI-DL \cite{gao2024harnessing}, a method that uses predicted gene expression from H\&E to predict patient prognosis. This is done by generating a graph, with predicted gene expression as nodes, to develop a survival model that outperforms others in breast and colorectal cancer cohorts.

\section{Morphology integration for spatial domain identification}

\begin{figure}[ht]
    \centering
    {\includegraphics[clip, trim=0cm 7cm 15.0cm 0cm, width=.65\linewidth]{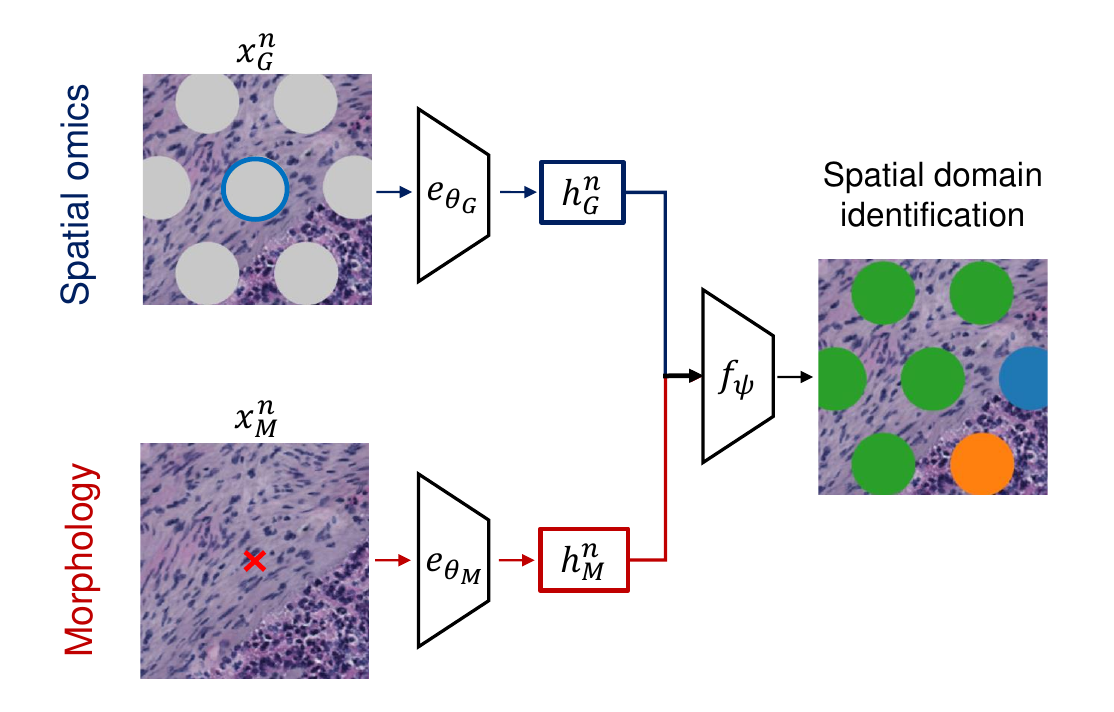}}
    \caption{The most common task involving morphology integration is the identification of spatial domains which can be then used in downstream tasks.}
    \label{fig:integration}
\end{figure}

Morphology integration involves identifying morphological features that spatially complement gene expression patterns. The primary application in this sceneario is spatial domain identification. The goal is to learn morphological features that are not correlated with gene expression but still add additional information to define meaningful regions (see Figure \ref{fig:integration}).

To the best of our knowledge, SpaCell \cite{tan2020spacell} is the first method that combines morphology with spatial gene expression. They propose a pretrained CNN to extract morphology that was then combined with gene expression to obtain a joint latent space. This integrated space was then used for clustering and domain identification, setting a foundational precedent for future studies.  

Hu \textit{et al.} proposed SpaGCN \cite{hu2021spagcn}, a graph-based method to further integrate spatial location with gene expression and histology. In this case, they use RGB intensity values as morphology descriptors and, through iterative clustering of the resulting graph they obtain spatial domains. 

The still unpublished conST \cite{zong2022const} proposes a contrastive approach for integrating gene expression, spatial information and morphology. The authors use a pretrained autoencoder to obtain histology representations which they then combine with gene expression in a graph neural network to obtain common features. The authors also implemented an interpretability module to explore the correlation between spots.

In a similar fashion as SpaCell, stLearn \cite{pham2023robust} also uses a pretrained CNN as a feature extractor. Together with the spot location, these features are used for normalization to adjust the gene expression values. They use this adjusted gene expression for tasks such as trajectory inference of cell-cell interaction analysis.

More recently, ConGI \cite{zeng2023identifying} proposed contrastive learning for obtaining a joint representation of histology and gene expression. The authors use a pretrained network as image encoder and three different contrastive losses to model the relationships between the two modalities.

\subsection{Extracting features that complement gene expression}

A good morphological feature extractor for integration ensures that the obtained features have complementary information to gene expression and do not add noise, so as to obtain more meaningful domains compared to using gene expression alone. 

Using CNN pre-trained on large datasets like ImageNet \cite{deng2009imagenet} is by far the most common approach to obtain morphological descriptors for integration and other tasks. Authors typically use the penultimate layer before prediction. It is the case for SpaCell and stLearn, which use a pretrained ResNet50 \cite{he2016deep} and ConGI, which uses DenseNet-121 \cite{huang2017densely}. conST employed an ImageNet pre-trained masked autoencoder (MAE) \cite{he2022masked} to obtain morphological features. In contrast, SpaGCN does not use any deep learning approach and instead extracts RGB values from the patches. 

As these features are not explicitly trained to capture gene expression patterns, one could argue that they probably fall in the left half (quadrants I and III) of Figure \ref{fig:intuition}. Authors probably expect that these features are general enough to add morphological information that is not already contained in the gene expression. But it is easy to see that networks trained on natural images or RGB values might not provide features which are relevant enough for the integration task \cite{chelebian2024self}. 

\subsection{Fusion modules for morphological and gene expression features}
Apart from the choice of image encoder, it is important to decide how to integrate the modalities: the fusion module. It is worth noting that all integration methods use modality specific encoders, due to the fixed workflow established for spatial omics, but a transformer-based architecture to handle input tokens for both modalities and use the same encoder for both is an interesting avenue \cite{zong2023self}.

\textbf{No explicit fusion.} SpaGCN does not have an explicit fusion module, instead, it constructs a graph with molecular features as the nodes and a combination of the spatial location and RGB image intensity features as the distance. stLearn also does not use a module for fusing the data and instead it uses the 50 principal components (PCs) from the 2048 features pretrained ResNet50 together with the spatial locations to normalize the gene expression.  

\textbf{Representation stitching.} SpaCell concatenates the latent representations of two auto-encoders that input the 2048 last layer features from the pretrained ResNet50 on the imaging side and the 2048 top highly variable genes from spatial transcriptomics.

\textbf{Representation fusion.} ConST generates a graph with KNN of spatial coordinates as edges and a concatenation of the 768 features extracted from the pretrained MAE and the 300 PCs from the genes as node features. It then generates the lower dimensional graph with them by self-supervised learning. ConGI also concatenates the representations from a gene expression MLP encoder and a DenseNet121 CNN for the images and feeds it into a small neural network to reduce the dimensionality in a self-supervised way as well.  

\section{Evaluation metrics, datasets and benchmarks}

\subsection{Metrics for assessing the learned representations}

\begin{figure}[ht]
    \centering
    {\includegraphics[clip, trim=0cm 0cm 2.7cm 0cm, width=\linewidth]{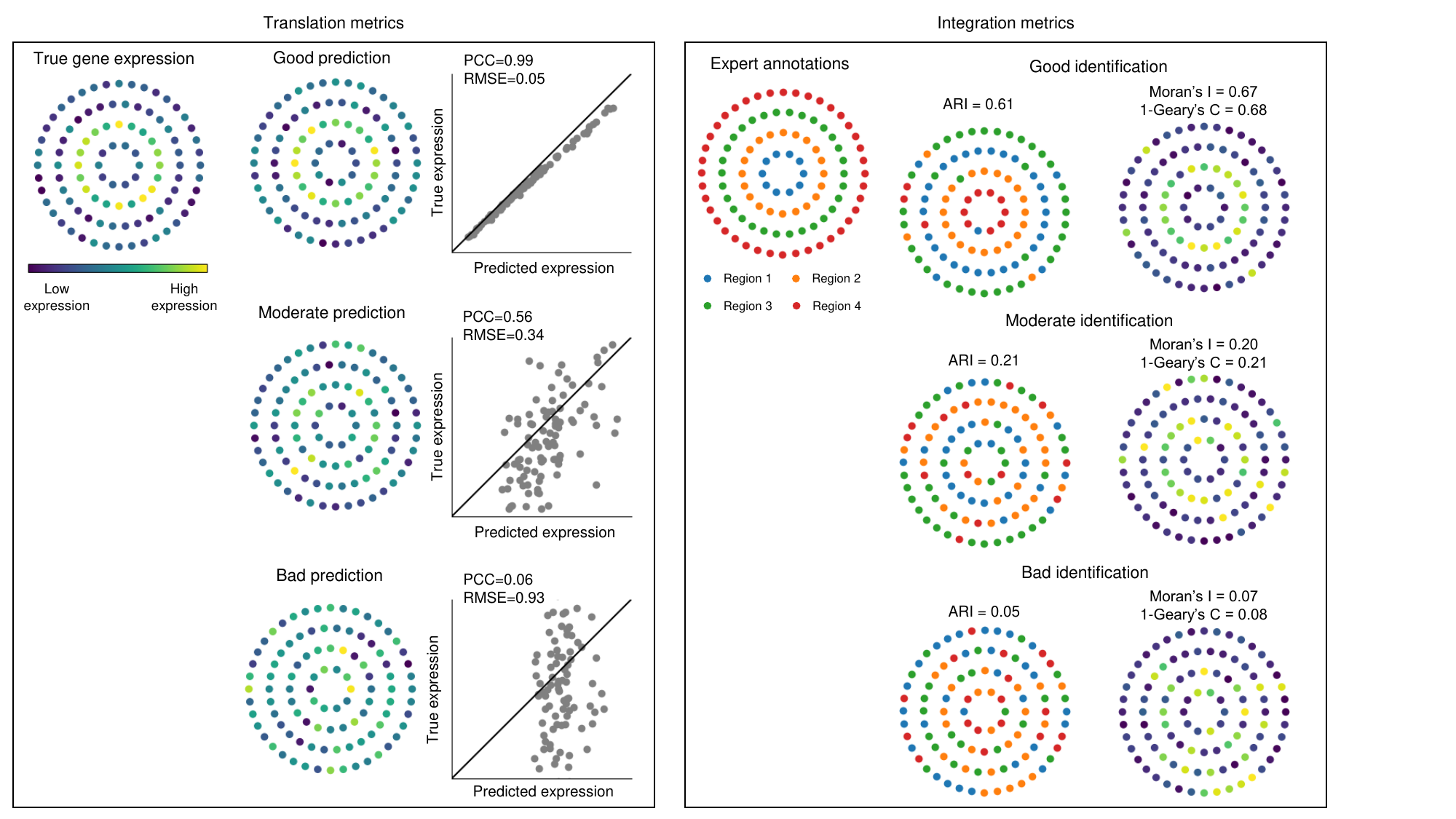}}
    \caption{Commonly used metrics for morphology translation and integration. This synthetic example presents tissue regions as concentric circles. (a) Translation metrics usually measure the agreement of the true gene expression with the gene expression predicted from morphology. Pearson's correlation coefficient (PCC) and regression metrics such as the root mean squared error (RMSE) are usually employed. (b) Integration metrics measure the agreement of expert annotations with the domains defined jointly morphology and spatial transcriptomics via the adjusted Rand index (ARI). It is common to also define spatially variable genes that define the identified domains and measure their degree of spatial auto-correlation with Moran's I or Geary's C.}
    \label{fig:metrics}
\end{figure}

Selecting the best possible representation of tissue morphology, both for translation and integration tasks, requires metrics to evaluate how well the learned features may solve the task at hand. The metrics introduced by the pioneering publications in the field are still widely used, despite their limitations \cite{chan2023benchmarking, yuan2024benchmarking}. Figure \ref{fig:metrics} shows an overview of these commonly used metrics.

\subsubsection{Metrics for assessing morphological translation}

In our framework, the ideal metric would measure how well the morphological features describe the gene patterns but, ideally, also include some information about their biological relevance. Authors have typically circumvented this by either selecting meaningful genes for prediction or analyzing the predictions \textit{ad hoc}. However, there is currently no way of assessing this in a quantitative manner, so many studies achieving high performance metrics might be in the overestimation quadrant of the framework, having good results on non-informative genes.

\textbf{Pearson's correlation coefficient}

The Pearson's correlation coefficient ($\text{PCC}$), also represented by $\rho$, is a measure of the linear relationships between the observed and the predicted gene expression. Mathematically, it is computed as the covariance ($\text{cov}$) of the two variables divided by the product of their standard deviations ($\sigma$). The covariance is a measure of the tendency between the observed and predicted expression i.e. if higher true expression levels match higher predicted levels, the covariance is higher. The division by their $\sigma$ is a way of normalizing so that the metric is bound between $+1$ perfect positive and $-1$ perfect negative correlation.

Let $y$ be the observed true gene expression for a gene, $\hat{y}$ be the predicted gene expression for that gene, and $\sigma_{y}$ and $\sigma_{\hat{y}}$ their standard deviations, then the PCC can be expressed by: 

\begin{equation}
    \text{PCC} = \frac{\text{cov}(y, \hat{y})}{\sigma_{y} \sigma_{\hat{y}}}
\end{equation}

\textbf{Regression metrics}

The problem of gene expression prediction is typically framed as a regression problem. More recent methods also report the performance of different regression metrics, namely mean absolute error (MAE), mean square error (MSE) and its root MSE (RMSE). Let \(y_i\) be the observed true gene expression for a gene at position $i$ and \(\hat{y}_i\) be the predicted gene expression for that gene at the same position. In other words, a small error means that we have learned features that predict gene expression well.

The MAE measures the average magnitude of the errors in a set of predictions, without considering their direction. It is the average over the test sample of the absolute differences between prediction and actual observation where all individual differences have equal weight.

\begin{equation}
    \text{MAE} = \frac{1}{n} \sum_{i=1}^{n} |y_i - \hat{y}_i|
\end{equation}

The MSE measures the average of the squares of the errors—that is, the average squared difference between the estimated values and the actual value. MSE is a risk function corresponding to the expected value of the squared error loss.

\begin{equation}
\text{MSE} = \frac{1}{n} \sum_{i=1}^{n} (y_i - \hat{y}_i)^2
\end{equation}

The RMSE is the square root of the average of squared differences between prediction and actual observation. It provides a measure of how spread out these residuals are. In other words, it tells you how concentrated the data is around the line of best fit.

\begin{equation}
\text{RMSE} = \sqrt{MSE} = \sqrt{\frac{1}{n} \sum_{i=1}^{n} (y_i - \hat{y}_i)^2}
\end{equation}

Each of these metrics offers a different perspective on the accuracy of the model predictions. MAE gives an idea of the magnitude of errors, MSE emphasizes larger errors more than smaller ones due to squaring the differences, and RMSE provides an overall measure of fit in the same units as the response variable. However, the values are not bounded, which can be harder to interpret and compare. 

\subsubsection{Metrics for assessing morphological integration}

In our framework, the ideal metric for identifying strong features for morphological integration would measure how the morphological features, despite not being correlated with gene expression patterns, are still relevant and thus add additional information to the spatial domain identification. On the one side, authors typically measure how well their clustered domains agree with expert annotations. On the other, the defined domains can be linked to a specific spatially variable gene (SVG). This gene can be validated by its amount of auto-correlation. 

\textbf{Clustering evaluation metrics}

Since spatial domain identification is not a classification problem in which there is a one-to-one connection between regions, we need metrics that take this into account. The Rand index (RI) and its adjusted version (ARI) \cite{rand1971objective, hubert1985comparing, vinh2009information} measure the similarity of two clusters, accommodating for scenarios with unlabeled and different amounts of categories. To evaluate the similarity between two clustering results, we use metrics that account for such complexities.

The ARI is a measure of the similarity between two data clusterings, correcting for the chance grouping of elements. The ARI ranges from -1 to 1, where 1 indicates perfect agreement between the two clusterings, 0 indicates random clustering, and negative values indicate less agreement than expected by chance.

\begin{equation}
    \text{ARI} = \frac{\text{RI} - \text{Expected RI}}{\max(\text{RI}) - \text{Expected RI}}
\end{equation}

where RI is the measure of the similarity between two data clusterings, and Expected RI is the expected value of the RI for random clustering.

\textbf{Auto-correlation metrics}

Gene expressions at different locations may exhibit spatial auto-correlation, where nearby locations have similar expression levels. To assess this, authors use Moran’s I \cite{moran1950notes, li2007beyond} and Geary’s C \cite{geary1954contiguity, anselin2019local} statistics.

Moran’s I quantifies overall spatial auto-correlation of gene expression. It ranges from -1 to 1, where 1 indicates a clear spatial pattern, 0 indicates random spatial expression, and -1 indicates a chessboard-like pattern.

\begin{equation}
\text{Moran's I} = \frac{N}{W} \frac{\sum_{i}\sum_{j}w_{ij}(x_i - \bar{x})(x_j - \bar{x})}{\sum_{i}(x_i - \bar{x})^2}
\end{equation}

where \(x_i\) and \(x_j\) are the gene expressions at spots \(i\) and \(j\), \(\bar{x}\) is the mean expression, \(N\) is the total number of spots, \(w_{ij}\) is the spatial weight between spots \(i\) and \(j\), and \(W\) is the sum of \(w_{ij}\). We set \(w_{ij} = 1\) for the 4 nearest neighbors of spot \(i\) and \(w_{ij} = 0\) otherwise.

Geary’s C also measures spatial autocorrelation, but focuses on local differences. Its value ranges from 0 to 2. To align it with Moran’s I, we scale it to [-1, 1]:

\begin{equation}
\text{Geary's C} = \frac{(N-1)}{2W} \frac{\sum_{i}\sum_{j}w_{ij}(x_i - x_j)^2}{\sum_{i}(x_i - \bar{x})^2}
\end{equation}

The scaled version is:

\begin{equation}
\text{Scaled Geary's C} = 1 - \text{Geary's C}
\end{equation}

Here, 1 indicates perfect positive autocorrelation, 0 indicates no autocorrelation, and -1 indicates perfect negative autocorrelation. Both metrics provide insight into the spatial patterns of gene expression.

\subsection{Public datasets and benchmarks}

As with the metrics, the datasets that the community has used for developing new methods draw inspiration from the first works. The scarce availability of big spatial transcriptomics datasets further enforced this.

Many translation methods methods use breast cancer datasets, either the two ductal carcinoma samples from 10X Visium and the 23 breast cancer patients used by ST-Net \cite{he2020integrating}, or the 32 HER2-positive breast cancer samples presented by Andersson \textit{et al.} \cite{andersson2020transcriptome}. Another commonly used dataset has been the human squamous cell carcinoma dataset \cite{ji2020multimodal}. This is also the case for the  benchmark presently available for translation studies \cite{chan2023benchmarking}. Even though the authors also include methods that do not train on spatial data, they analyze the performance of ST-Net, DeepSpaCE and HisToGene on the HER2-positive and squamous cell carcinoma datasets.

Super-resolution tasks are typically harder to evaluate quantitatively, as there is no ground-truth for per-pixel gene expression. New technologies such as 10X Xenium were utilized as a surrogate for this by iStar and MHAST. This enabled the authors to have a denser gene expression map that could be then compared to their methods, even though 10X Xenium is a imaging-based method and thus requires predefining a gene panel of 100s of genes, far from the whole transcriptome.

The two main datasets used for integration methods are 10X Visium mouse brain samples and the 10X Visium human dorsolateral prefrontal cortex available at the spatialLIBD library \cite{pardo2022spatiallibd}.  The latter was used in the benchmark by Yuan \textit{et al.} \cite{yuan2024benchmarking}. As the integration tasks is commonly spatial domain identification, it requires some type of annotations for comparison, limiting the amount of available datasets.

\section{Technical considerations}

Working with histological images comes with its own set of challenges that are far from solved \cite{tizhoosh2018artificial, bera2019artificial, van2021deep}. One of the primary issues is the inherent variability in histological image data, which arises from differences in staining protocols, sample preparation techniques, and imaging conditions across different laboratories and studies. This variability complicates the generalization of models, as algorithms trained on one dataset may not perform well on another due to these inconsistencies. Pretraining models on a wide variety of data from different sources can help improve their robustness and generalization capabilities \cite{xu2024whole}.

To this we add the challenges from working with such high dimensional data as spatial transcriptomics. Each modality individually produces complex, high-dimensional data, and integrating these datasets compounds the complexity. Dimensionality reduction techniques are usually employed to reduce the complexity of the data while trying to retain the essential information \cite{shang2022spatially, sun2024comprehensive}. 

These challenges make it hard to benchmark and compare methods. Many authors compare against previous work consistently outperforming them, but this is done in new datasets and with different validations schemes. The low availability of big datasets limits the generalization of current benchmarks, at times yielding contradictory results with the original authors \cite{chan2023benchmarking, yuan2024benchmarking}. Bigger annotated datasets are necessary for method development and comparison. Transparent reporting of training and validation procedures is also crucial to ensure reproducibility and allow for meaningful comparisons across studies.

An important consideration for our study is the reassessment of the metrics currently utilized, with the possibility of incorporating an additional dimension. Traditional translation tasks often employ PCC or regression metrics, which primarily evaluate the shared information between learned morphological features and gene expression patterns, analogous to the x-axis in our proposed framework. This approach can lead to an overestimation of predictive performance for genes that are not task-relevant, such as housekeeping or constitutive genes, neglecting genes that may hold significant clinical value. Similarly, integration methods typically assess the shared information between the fusion of morphological features and gene expression with expert annotations, comparable to the y-axis in our framework. This methodology can entangle the individual contributions of each modality and potentially diminish performance by incorporating noisy or correlated morphological features.

Finally, an often overlooked aspect is the extensive training and computational resources required in these applications. In predicting gene expression, especially with internal validation as opposed to an independent test set, additional training can substantially enhance results. This concern also extends to integration tasks, where it becomes difficult to determine whether improved performance is attributable to the actual contribution of morphological data or merely the result of overfitting to the specific problem.

\section{Outlook and perspectives}

The combination of morphology and spatial transcriptomics is at its dawn. In this work we presented a mental framework to conceptualize and guide future developments in this field, as many challenges remain unresolved.

Correlation values for gene expression prediction are rather moderate. Even with modern implementations, performances are far from clinically transferable. If we want to be able to apply such methods to predict the gene expression in large H\&E-stained cohorts, we have to make sure we are falling in the first quadrant of Figure \ref{fig:intuition}. Achieving this will require not just an improvement in average model performance, but a focus on ensuring that methods are truly effective for clinically important genes. This might involve developing specialized models tailored to specific clinical problems, ensuring their applicability in real-world scenarios.

The task of spatial domain identification has not yet shown significant benefits from morphological integration. This is likely due to the fact that the morphological features currently used are either not relevant enough or provide redundant information already captured by gene expression data. Future method development should prioritize the creation of morphological descriptors that are both highly relevant and complementary to gene expression, ensuring that they add value rather than noise to the analysis. Until such improvements are made, the integration of morphology should be approached with caution, avoiding the inclusion of irrelevant or redundant features.

The framework proposed in this review is not limited to the integration of morphology and spatial transcriptomics; it can be extended to other multi-omics or multi-modal data integration tasks. In cases of translation, the goal is to maximize the shared information between modalities while focusing on the specific downstream task. For integration, the aim is to minimize redundant information while ensuring that both modalities contribute meaningfully to the final analysis. This flexible framework can guide future research in multi-modal data integration, ensuring a more systematic and effective approach.

Ultimately, we hope this framework will serve as a compass for researchers, facilitating the joint use of morphology and spatial transcriptomics as imaging AI and bioinformatics continue to advance. By providing a structured approach to integrating these complex data types, this framework aims to guide efforts to gain deeper insights into biological processes and disease mechanisms.

\section*{Acknowledgments}
We would like to thank current and former collaborators and co-workers in the field of spatial omics and imaging AI for sharing their insights leading up to this review. 

\printbibliography

\end{document}